\documentclass[letterpaper, 10 pt, conference]{ieeeconf}  

\IEEEoverridecommandlockouts                              

\usepackage{graphics} 
\usepackage{epsfig} 
\usepackage{mathptmx} 
\usepackage{times} 
\usepackage{amsmath} 
\usepackage{amssymb}  
\usepackage[T1]{fontenc}
\usepackage{caption}
\usepackage{graphicx}
\usepackage{multirow}
\usepackage{tabularx}
\usepackage{pifont}
\usepackage{multirow}
\usepackage{float}
\usepackage{booktabs} 
\usepackage{pifont}
\usepackage{hyperref} 
\usepackage{fontawesome}

\title{\LARGE \bf
EmbodiedCoder: Parameterized Embodied Mobile Manipulation via \\ Modern Coding Model
}

\author{
  \textbf{Zefu Lin}$^{2,3}$\quad
  \textbf{Rongxu Cui}$^{5}$\quad
  \textbf{Chen Hanning}$^{1}$\quad
  \textbf{Xiangyu Wang}$^{1,2,3}$\quad
  \textbf{Junjia Xu}$^{5}$\quad
  \textbf{Xiaojuan Jin}$^{2,3}$\\
  \textbf{Chen Wenbo}$^{2,3}$ \quad
  \textbf{Hui Zhou}$^{6}$ \quad
  \textbf{Lue Fan}$^{2,3 }$ \textsuperscript{\faEnvelopeO}\quad
  \textbf{Wenling Li}$^{5}$ \quad
  \textbf{Zhaoxiang Zhang}$^{1,2,3,4 }$ \textsuperscript{\faEnvelopeO}
  \\
$^1$ University of Chinese Academy of Sciences (UCAS)\\
$^2$ Institute of Automation, Chinese Academy of Sciences (CASIA)\\
$^3$ New Laboratory of Pattern Recognition (NLPR)\\
$^4$ State Key Laboratory of Multimodal Artificial Intelligence Systems (MAIS)\\
$^5$ Beihang University \quad
$^6$ Chinese University of Hong Kong \\
  \tt\small{\{linzefu2022, lue.fan\}@ia.ac.cn}  \quad 
}

\begin{document}

\thispagestyle{empty}
\pagestyle{empty}


\twocolumn[{%
\renewcommand\twocolumn[1][]{#1}%
\maketitle
\vspace{-10mm}
\begin{center}
    \captionsetup{type=figure}
    \includegraphics[width=1\linewidth]{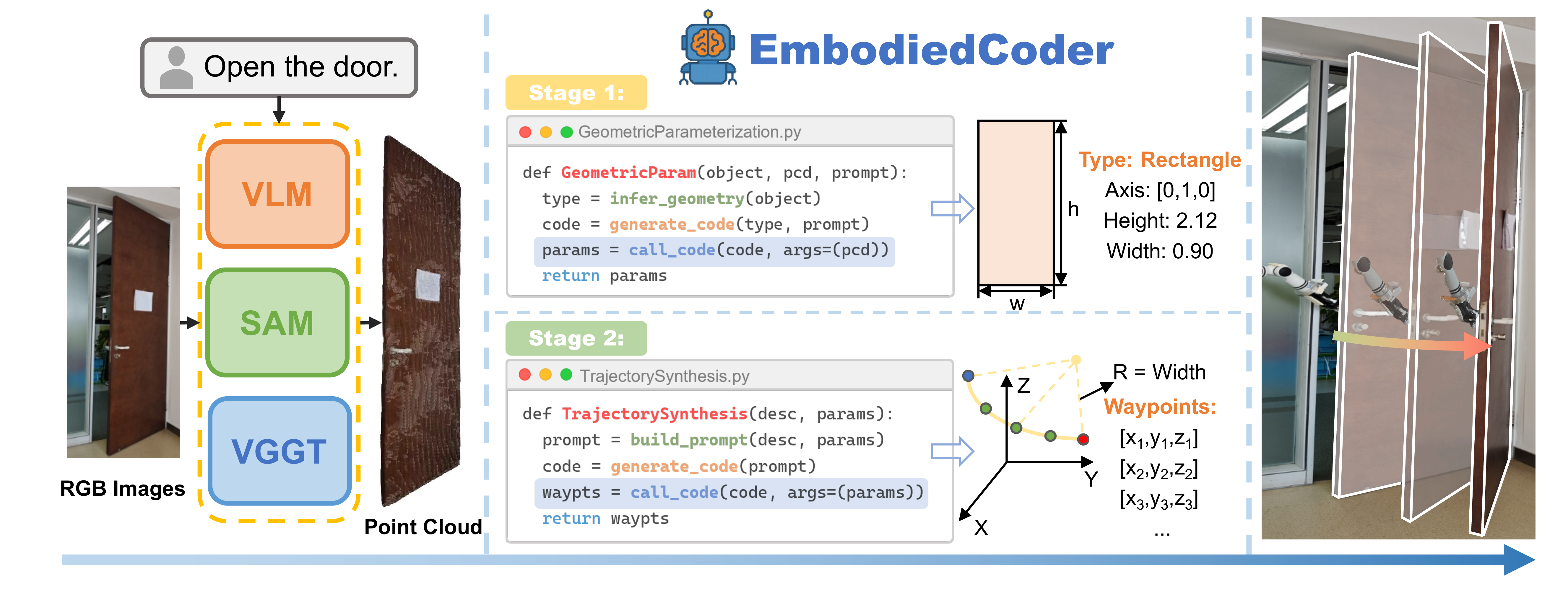}
    \vspace{-8mm}
    \captionof{figure}{    
    \textbf{EmbodiedCoder} employs code generation to bridge perception and manipulation by parameterizing objects and synthesizing task-specific trajectories. The figure shows a subtask \textbf{\textit{Open the door}} derived from a long-term instruction. Through code-driven geometric parameterization, the door is represented as a parametric model with a hinge axis, and the system generates code that synthesizes a semicircular trajectory consistent with this geometry. The robotic arm then executes the opening motion by following waypoints sampled from the generated trajectory, demonstrating how coding enables functional manipulation without additional training.
    }\label{fig: highlight}
    
\end{center}%
}]

\begin{abstract}
Recent advances in control robot methods, from end-to-end vision-language-action frameworks to modular systems with predefined primitives, have advanced robots’ ability to follow natural language instructions. Nonetheless, many approaches still struggle to scale to diverse environments, as they often rely on large annotated datasets and offer limited interpretability.
In this work, we introduce \textit{EmbodiedCoder}, a training-free framework for open-world mobile robot manipulation that leverages coding models to directly generate executable robot trajectories.
By grounding high-level instructions in code, \textit{EmbodiedCoder} enables flexible object geometry parameterization and manipulation trajectory synthesis without additional data collection or fine-tuning.
This coding-based paradigm provides a transparent and generalizable way to connect perception with manipulation. 
Experiments on real mobile robots show that \textit{EmbodiedCoder} achieves robust performance across diverse long-term tasks and generalizes effectively to novel objects and environments. 
Our results demonstrate an interpretable approach for bridging high-level reasoning and low-level control, moving beyond fixed primitives toward versatile robot intelligence. See the project page at \href{https://embodiedcoder.github.io/EmbodiedCoder/}{https://embodiedcoder.github.io/EmbodiedCoder/}.
\end{abstract}

\section{INTRODUCTION}

Enabling robots to perform diverse tasks with human-like proficiency in complex, unstructured environments has long been a central goal in robotics~\cite{kaelbling2020foundation}. 
Recent progress in vision-language-action (VLA) models has brought this ambition closer to reality by enabling end-to-end mapping from sensory inputs and natural language instructions to robot actions. However, their generalization ability remains limited. Even slight changes in the environment, such as variations in object appearance or illumination, can significantly degrade performance. Furthermore, these models typically require massive annotated datasets, making their deployment costly and less scalable.

To overcome these challenges, hierarchical strategies have been proposed. A common design, such as DovSG~\cite{dovsg} and OK-Robot~\cite{okrobot}, is to employ vision-language models (VLMs)~\cite{clip, li2023blip} as high-level planners that decompose tasks into subtasks and invoke predefined robotic primitives, such as navigation, grasping, or pick-and-place. This paradigm is theoretically appealing, since it allows robots to leverage the commonsense knowledge encoded in large-scale models while relying on robust control modules for low-level execution. In practice, however, its effectiveness is fundamentally constrained by the repertoire of available manipulation primitives. Many real-world tasks, such as opening doors or drawers, require nuanced interactions that cannot be reduced to a finite set of predefined primitives.

Recent work has attempted to extend beyond this primitive-based architecture by generating executable code for manipulation. Code-as-Policies~\cite{Codeaspolicies} demonstrated that an LLM can write low-level code to control a robot, but this early attempt was limited to tasks with very simple, specific geometries. RoboCodeX~\cite{mu2024robocodex} uses a multimodal code generation framework to broaden task generality, yet it relies on learned models to handle physical constraints, which reduces its adaptability to novel scenarios. VoxPoser~\cite{huang2023voxposer} computes obstacle-aware end-effector trajectories for tasks like drawer opening, but it cannot perform more intricate, contact-rich manipulations. Likewise, Code-as-Monitor~\cite{codeasmonitor} generates code to detect and recover from execution failures, but it does not expand the robot’s basic manipulation repertoire beyond the original primitives.
For wheeled robots, the task complexity becomes even higher~\cite{16}. The robot must be able to retain information about the environment, which allows it to incorporate objects beyond its immediate field of view into the task planning process.

To address these challenges, we propose \textbf{EmbodiedCoder}, a code-driven framework for open-world mobile robot manipulation. 
Unlike traditional training-intensive approaches, our method leverages the expressive power of coding models to generate executable code that directly encodes manipulation strategies. 
This design transforms high-level instructions into programmatic representations of geometric parameterization and trajectory synthesis. 
By grounding the reasoning process in code, the system benefits from both interpretability and flexibility, enabling robots to adapt to novel objects and environments without additional training or fine-tuning.

At the core of our framework, code serves as the medium that bridges perception and manipulation. 
The process begins with scene understanding, where VGGT~\cite{wang2025vggt} and a vision-language model capture RGB-D observations and ground semantic information into 3D point representations. 
Based on this input, \textbf{EmbodiedCoder} prompts the coding models to generate code for two critical stages. 
First, in \textit{code-driven geometric parameterization}, the system fits point clouds of task-relevant objects to geometry parametric primitives that encode functional affordances, such as approximating a drawer as a cuboid with a pulling axis. 
Second, in \textit{code-driven trajectory synthesis}, our method produces programmatic descriptions of feasible motion trajectories that satisfy physical, environmental, and task-specific constraints. 
The trajectories are first represented as parameterized curves, from which discrete waypoints are sampled and subsequently executed by the robot.
By this approach, the system not only achieves robust performance in novel environments but also provides a transparent and generalizable mechanism for linking perception with real-world manipulation.

In summary, our method not only alleviates the dependency on predefined primitives but also eliminates costly data collection and fine-tuning. 
Our contributions are threefold:
\begin{itemize}
    \item We introduce a framework that integrates coding models with embodied agents, enabling complex long-term manipulations in real-world environments.
    \item We propose a novel method for parameterizing objects into functional geometric abstractions, allowing pretrained knowledge to be grounded into executable trajectories for sophisticated.
    \item We validate EmbodiedCoder on real mobile robots and demonstrate its effectiveness in handling diverse tasks, showing improved generalization and training-free deployment compared to existing approaches.
\end{itemize}

\section{RELATED WORKS}

\subsection{Data-Driven Robotic Policies}
Vision-Language-Action (VLA) models map visual observations and instructions directly to low-level robot actions via large transformer policies~\cite{1,2,3,4,5}. RT-2~\cite{rt2}, for example, extends multi-task policies with web-scale vision-language pretraining to enhance zero-shot understanding. However, such models require massive robot demonstration datasets~\cite{rt1} and often fail to generalize under distribution shifts in lighting, appearance, or object variation. Recent efforts improve efficiency~\cite{6,7,8,9}. TinyVLA~\cite{wen2025tinyvla} achieves faster inference and greater data efficiency with a compact design, while GR00T N1~\cite{GR00T} employs dual-system reasoning for humanoid control. These advances broaden VLA capabilities, but even state-of-the-art policies still depend on curated datasets and lack systematic generalization in novel scenarios~\cite{10,11,12}.

\subsection{LLM-Driven Trajectory and Code Generation}
Another line of research employs large language models (LLMs) as high-level planners or code generators for robotics~\cite{16,17}. Code-as-Policies~\cite{Codeaspolicies} showed that LLMs can compose robot-executable code from natural language, but flexibility is limited by fixed APIs. VoxPoser~\cite{huang2023voxposer} generates 3D affordance maps for motion planning and zero-shot execution of tasks like drawer opening, though only for relatively simple manipulations. ReKep~\cite{huang2024rekep} detects object keypoints and prompts LLMs to produce relational cost functions, enabling multi-stage manipulation without task-specific training, though tasks poorly described by sparse keypoints remain difficult. RoboCodeX~\cite{mu2024robocodex} decomposes high-level instructions into object-centric units with affordances and safety constraints, generating structured code that achieves strong results across simulation and real robots. Yet, its reliance on curated multimodal data restricts adaptability. Complementary to planning, Code-as-Monitor~\cite{codeasmonitor} compiles natural language constraints into runtime monitors for failure detection and recovery. In summary, these LLM-driven methods~\cite{13,14,15} illustrate the potential of language models for behavior composition and robustness, but they still struggle with contact-rich interactions and depend heavily on prior data or predefined skills.

\subsection{Modular Systems with Predefined Skills}
A third strategy integrates high-level reasoning with a fixed library of skills. SayCan~\cite{saycan} combines language models with affordance-based value functions to choose among predefined behaviors, producing interpretable long-term plans but constrained by a finite repertoire. OK-Robot~\cite{okrobot} integrates open-vocabulary object recognition with grasping and navigation, achieving strong zero-shot pick-and-place performance without extra training. DovSG~\cite{dovsg} leverages dynamic open-vocabulary 3D scene graphs to update the world model during execution, supporting adaptive planning in changing environments. Despite robustness and interpretability, these modular systems remain fundamentally bounded by predefined skills, limiting their ability to handle novel behaviors or tool use without manual extension.

In contrast, we introduces \textbf{EmbodiedCoder}, a training-free, code-driven framework that bridges perception and action by generating executable programs for geometric parameterization and trajectory synthesis, thereby overcoming the limitations of data-hungry policies, predefined skill libraries, and narrow keypoint-based reasoning.

\section{METHOD} 
\begin{figure*}[h]
    \centering
    \includegraphics[width=0.95\textwidth]{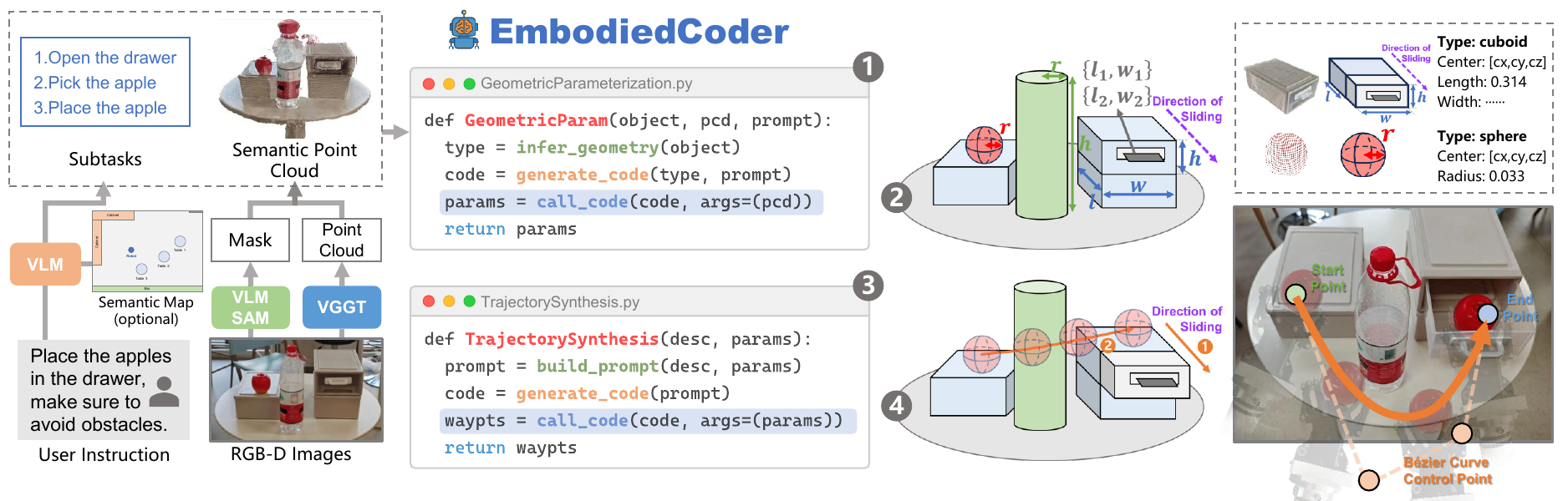}
    \caption{\textbf{Overview of the proposed system pipeline.} The system consists of three modules: (i) \textbf{\textit{Scene understanding and task decomposition}}, which processes RGB-D images with VLM and VGGT to build semantic maps and decompose instructions into subtasks; (ii) \textbf{\textit{EmbodiedCoder}}, which prompts an coding model to perform code-driven geometric parameterization of objects and trajectory synthesis under physical and environmental constraints; and (iii) \textbf{\textit{Motion execution}}, which samples waypoints from the synthesized trajectory and executes the manipulation with the robot arm.}
    \label{fig:pipeline}
    \vspace{-2.0em}
\end{figure*}

\subsection{Problem Setup}
We consider a mobile manipulation robot in everyday environments, tasked with executing complex instructions involving both navigation and manipulation. 
The robot must plan a sequence of actions $[a_1, \dots, a_N]$ that carries it from an initial state to a goal state satisfying the given instruction. 
We formalize the objective as a constrained motion planning process
\begin{equation}
F: (I_{rgbd}, L, C) \rightarrow [a_1, \dots, a_N],
\end{equation}
where $I_{rgbd}$ represents the RGB-D observations of the scene, $L$ is the natural language instruction, and $C$ denotes the set of constraints, including physical constraints of objects, environmental obstacles, and kinematic limits of the robot. 
The output is a sequence of actions aligned with the subtasks extracted from the instruction.

Rather than relying on a predefined library of low-level primitives whose limited expressiveness constrains manipulation generalization, we employ coding models to leverage their commonsense knowledge and strong code generation ability for object parameterization and trajectory resolution.
These models can qualitatively determine how objects should be parameterized and which trajectories are appropriate for accomplishing a given task, and they further provide quantitative solutions that transform these insights into executable motion plans.
Building on this capability, we propose \textbf{EmbodiedCoder}, a zero-shot framework that integrates coding models with robotic systems to plan and execute such tasks without additional training.

\subsection{System Overview}
The proposed system consists of three main modules as shown in Fig.~\ref{fig:pipeline}: 
(i) \textbf{Scene Understanding and Task Decomposition}. The module takes RGB-D images and task instructions as input, performs semantic grounding of objects, decomposes the instruction into subtasks, and outputs task-related semantic point clouds.
(ii) \textbf{EmbodiedCoder}. This core module generates the robot’s operation trajectories. Given a subtask and a semantic point cloud, EmbodiedCoder prompts the coding models to first parameterize objects by fitting point clouds to geometric primitives, and then to synthesize trajectories that conform to object geometry. Reasoning over object geometry allows the system to capture contact surfaces, spatial relations, and kinematic feasibility, which in turn ensures that the resulting trajectories satisfy environmental and physical constraints.
(iii) \textbf{Motion Execution}. This module carries out the planned motion by navigating to the target location and executing task-oriented manipulation.
The comparison with other methods is shown in the Table~\ref{tab:method_comparison}. We next describe each component in detail.
\newcommand{\cmark}{\ding{51}} 
\newcommand{\xmark}{\ding{55}} 

\begin{table}[!h]
\vspace{-0.5em}
\setlength{\abovecaptionskip}{-2pt}
  \centering
  \caption{Comparison of code-generation methods.}

  \label{tab:method_comparison}
  \setlength{\tabcolsep}{1pt}
  \renewcommand{\arraystretch}{1.15}
  \resizebox{\linewidth}{!}{
  \begin{tabularx}{\columnwidth}{l c c c c}
    \toprule
   \multirow{2}{*}{\textbf{Method}} & \textbf{Training-} & \multirow{2}{*}{\textbf{Code Type}} & \textbf{Skill} & \textbf{Long-term} \\ 

 & \textbf{Free} && \textbf{Library} & \textbf{Task} \\

    \midrule
    Code as Policies~\cite{Codeaspolicies}  &  \cmark & motion planing          & \cmark &  \\
    Code as Monitor~\cite{codeasmonitor}   &  & constraints             & \cmark &   \\
    RoboCodeX~\cite{mu2024robocodex}         &  & motion planing          & \cmark &  \\
    VoxPoser~\cite{huang2023voxposer}          &  & voxel value map         &   &  \\
    ReKep~\cite{huang2024rekep}             &  & constraints             &   &   \\
    CodeDiffuser~\cite{yin2025codediffuser} &  & perception &  &  \\
    RoboScript~\cite{chen2024roboscript} & \cmark & motion planing & \cmark &  \\
    \textbf{Ours}     & \cmark  & geometry \& trajectory &   & \cmark \\
    \bottomrule
  \end{tabularx}}
 
\end{table}
 \vspace{-1.5em}

\subsection{Semantic Scene Understanding and Task Decomposition}
\subsubsection{Semantic Mapping for Scene Understanding}
In the preparation phase, the robot processes a sequence of RGB images using VGGT~\cite{wang2025vggt} to reconstruct a dense point cloud of the entire scene. Relying solely on the RGB-D camera is insufficient due to depth noise and range limitations, so we align its depth maps with the VGGT reconstruction to achieve a reliable metric-scale point cloud. 
A vision-language model (VLM) provides semantic grounding of the scene in the form of bounding boxes, which are then passed to SAM~\cite{sam} to generate 2D semantic masks of all objects. These masks are projected onto the reconstructed point cloud, resulting in a semantic point cloud map. To facilitate subsequent task planning, this map is converted into a bird’s-eye-view semantic representation stored in 2D image form.

\subsubsection{Task Decomposition and Object-centric Semantic Understanding}
\label{subsubsec:task_decomposition}
In this stage, the system takes RGB-D observations and a complete task instruction as input. The instruction, together with the semantic map, is processed by the VLM to decompose the task into a sequence of subtasks, each associated with specific objects. For instance, the system may deduce that interacting with a door is a prerequisite before moving into an adjacent room.
For each subtask, the current RGB observation is fed into the VLM to identify and ground the relevant objects. 
In addition, the VLM infers the most suitable geometric shape of the objects and performs functional reasoning to determine which object parts must be manipulated to achieve the intended function.
The grounding results are passed to SAM~\cite{sam} to generate 2D semantic masks of these objects. The masks are then projected into the point cloud, producing a semantic point cloud representation that captures only the objects of interest for the given subtask. This object-centric representation is stored and later consumed by EmbodiedCoder for geometric parameterization and trajectory synthesis.

\subsection{EmbodiedCoder}
A central component of our framework is the EmbodiedCoder module, which leverages pre-trained large language models (LLMs) to transfer commonsense knowledge of manipulation into executable robot actions. This process is divided into two main stages.
  
\begin{figure}[!h]
    \centering
    \includegraphics[width=1\linewidth]{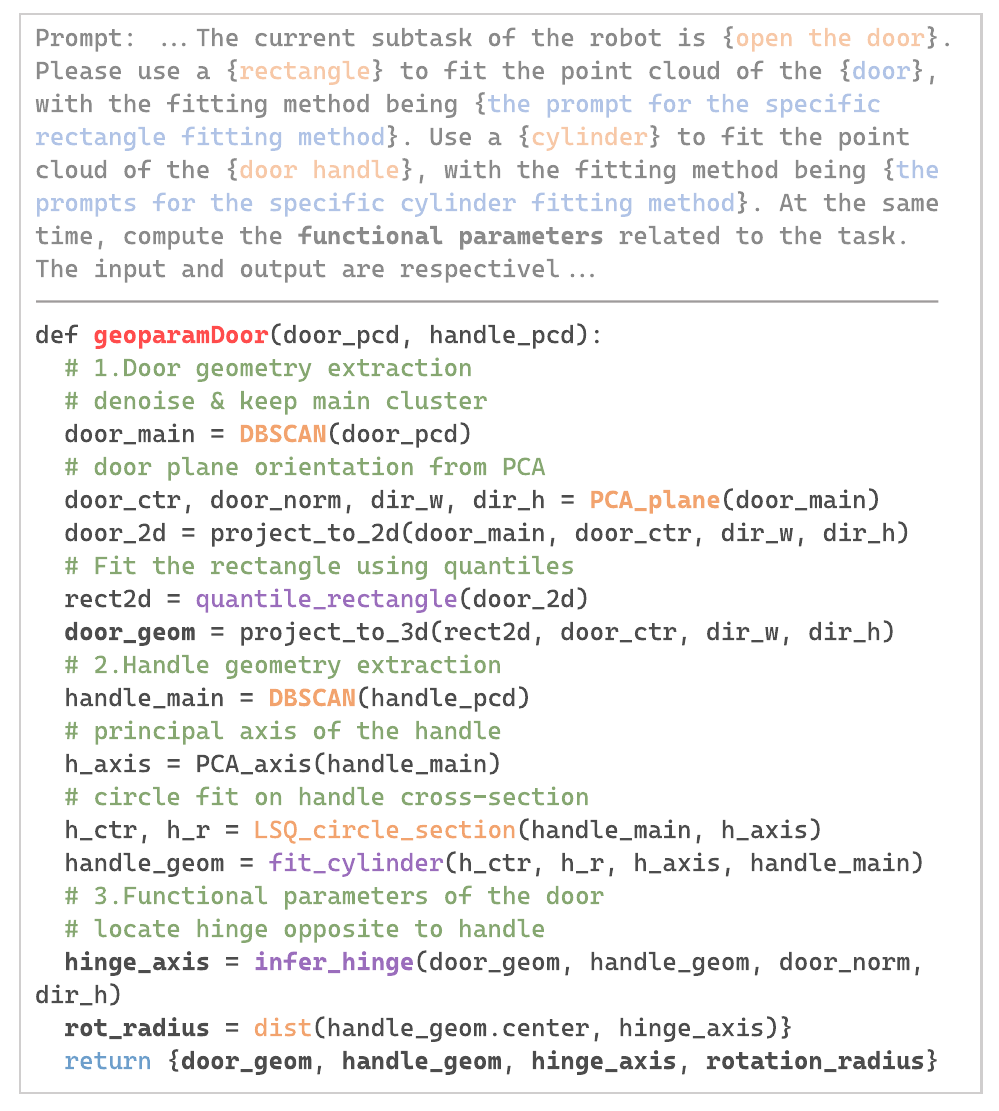}
    \vspace{-1.5em}
    \caption{\textbf{Example of door parameterization.} The task-relevant content in prompt is from \textit{Task Decomposition and Object-centric Semantic Understanding}.}
    \label{fig:code_for_para}
    \vspace{-2em}
\end{figure}

\textbf{(a) Code-driven Geometric Parameterization}.  
In this step, the system takes the task-relevant objects and their point clouds, and prompts the coding model to generate code that fits them to the geometric primitives identified in \textit{Task Decomposition and Object-centric Semantic Understanding}.
For instance, fitting a cylinder requires estimating its radius, height, and center position, while fitting a cuboid requires determining the length, width, height, and centroid coordinates. 
This process transforms incomplete or occluded point clouds into compact geometric parameterizations, facilitating more robust reasoning. 
For deformable objects, extreme points are selected to construct bounding envelopes instead of rigid parameterizations.
Fig.~\ref{fig:code_for_para} presents the code-driven parameterization of a door.

\begin{figure*}[th]
    \centering
    \includegraphics[width=0.9\textwidth]{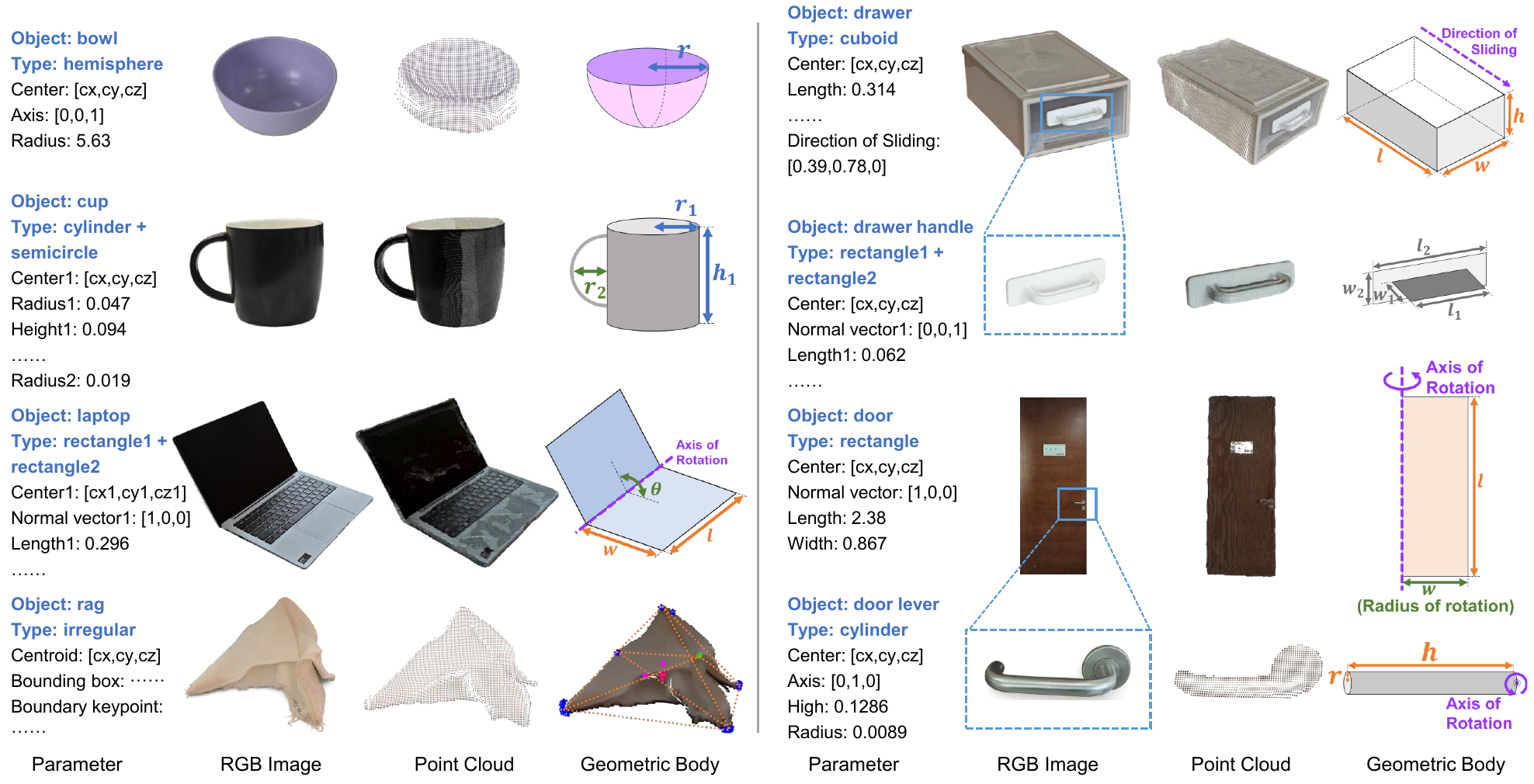}
    \caption{Examples of Parameterization Result for Common Objects.}
    \label{fig:obj geometry}
    \vspace{-2.0em}
\end{figure*}

Extracting geometric parameters for all task-relevant objects provides a structured basis for trajectory planning, enabling the system to satisfy constraints such as obstacle avoidance and motion feasibility. 
Consequently, we parameterize not only the overall structure but also the functional components that support interaction, such as drawer handles or door knobs.
This process directs the robot’s attention to task-relevant regions of the object and leads to more accurate and effective manipulation.
After this step, unstructured point cloud data are converted into structured representations. For example, the point cloud of an apple is reduced to a sphere defined by its center and radius, while a door is represented as a combination of a cuboid for the panel, a rotational axis, and a cylinder for the handle. Such compact parameterizations make the subsequent generation of task-specific trajectory code feasible.
Fig.~\ref{fig:obj geometry} shows some visualization results.

\textbf{(b) Code-driven Trajectory Synthesis}.  
After obtaining the geometric parameters of a task-specific object, we prompt the coding model to generate code for synthesizing a trajectory   that aligns with the object’s functional properties and the task requirements.

The trajectory generation process accounts for multiple constraints, which are inferred directly by the coding model. 
For example, in the case of opening a door, the physical constraint is that the door must rotate around its hinge axis; 
the environmental constraint concerns whether the door should be pushed or pulled; and the hardware constraint of the robot arises from the limited range of motion of its joints.
Additionally, the door’s opening gap must be wide enough to allow the robot to pass through, not just a small opening. This requires taking the robot’s dimensions into consideration when planning the trajectory.
In environments with obstacles, the system parametrizes the obstacles and incorporates them into the trajectory planning process. 
This ensures that the generated trajectory avoids these obstacles, adapting the robot’s movements to navigate around them effectively.
Coding model generates the appropriate trajectory generation code based on the task requirements and the object’s parameters, enabling the system to generalize across various tasks without manual design for each individual scenario.

\begin{figure}[!h]
    \centering
    \vspace{1em}
    \includegraphics[width=1\linewidth]{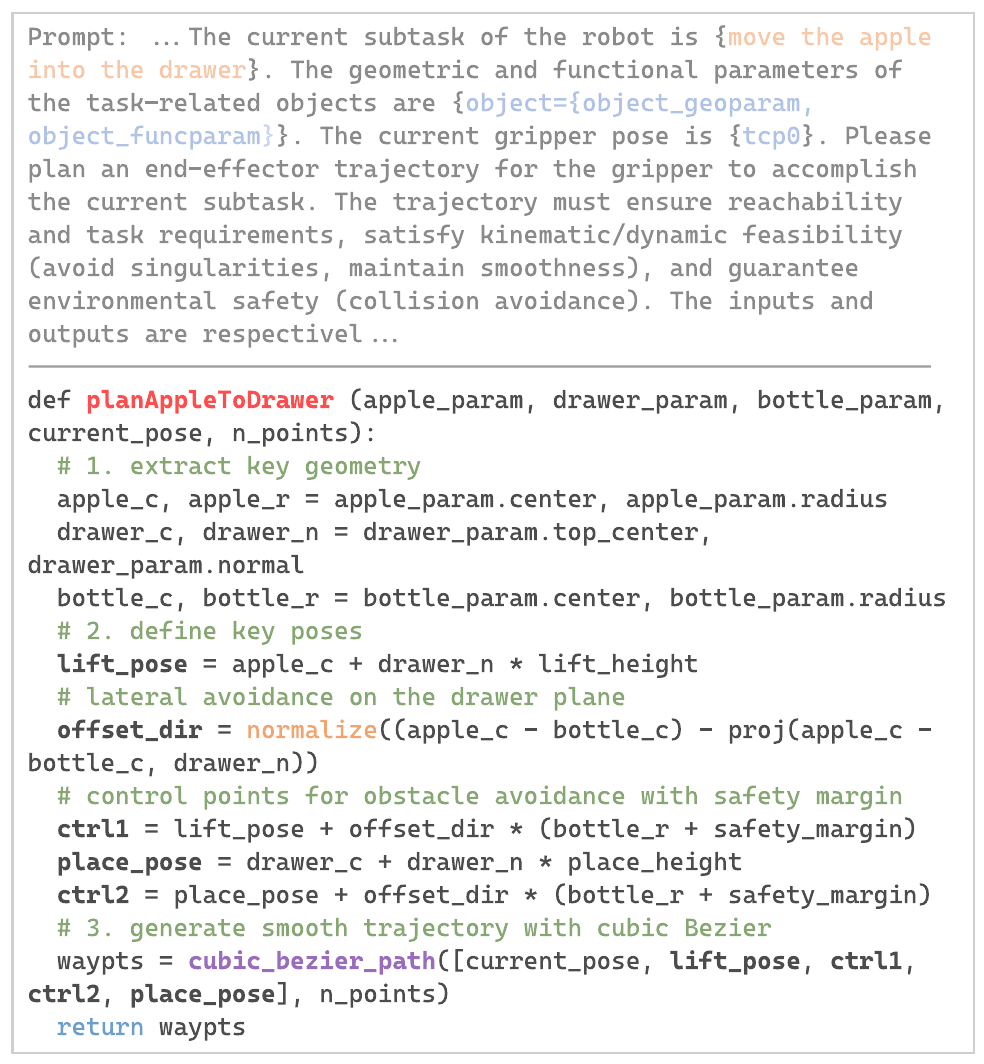}
    \vspace{-2em}
    \caption{\textbf{Example of apple placement with obstacle avoidance}. The task-relevant content in prompt is obtained through subtask decomposition and geometric parameterization.}
    \label{fig:code_for_curve}
    \vspace{-1em}
\end{figure}

Once the trajectory form is determined, it is expressed as a parametric curve, such as a line, arc, or Bézier curve, with the necessary parameters for the specific task.
Fig.~\ref{fig:code_for_curve} shows the code generated from the prompt for placing an apple while avoiding obstacles.
Using code generation for trajectory synthesis, instead of relying on a VLA that directly maps visual inputs to actions, improves the interpretability of the process.
This method provides the flexibility to dynamically adjust the trajectory in response to changing task conditions or environments, ensuring that the robot can generalize to new, previously novel tasks.

\textbf{(c) Code Caching}.  
For tasks involving familiar object types or recurring subtasks, the system reuses previously generated code, which can reduce latency and prevent code generation failures caused by unsuccessful model reasoning.
For novel objects or novel tasks, EmbodiedCoder is invoked to parameterize the object and synthesize new trajectories. 
As the system successfully executes more tasks, it gradually builds a growing library of versatile skills that can be applied to future problems.
This design achieves a balance between efficiency for known cases and generalization for open-world scenarios.

\subsection{Motion Execution}
After scene understanding and task decomposition, the robot executes each subtask sequentially. EmbodiedCoder generates a trajectory for each subtask, from which the robot samples waypoints to navigate and perform the required manipulation.
At this point, the entire process from vision-language input to action output has been completed.

\section{EXPERIMENT}
We evaluate the proposed system in real-world environments on long-term mobile-manipulation tasks. The experiments test (i) whether code-driven geometric parameterization and trajectory synthesis translate language goals into executable motions, (ii) how well the system generalizes to novel tasks compared with VLA and other code-generation methods, and (iii) the contribution of each module through ablations.

\subsection{Experimental Setup}
All experiments are conducted on an AgileX Cobot S Kit with a RealSense D455 RGB-D camera. The Scene Understanding and Task Decomposition module uses Qwen-2.5-VL~\cite{qwen} (7B) for grounding and instruction decomposition, SAM~\cite{sam} for masks, and VGGT~\cite{wang2025vggt} for reconstruction to metric point clouds. EmbodiedCoder employs Claude-Sonnet-4~\cite{claude} to generate parameter-fitting and trajectory synthesis code. Unless otherwise stated, these models are used throughout; ablations replace either the VLM or the coding model to assess sensitivity.

\subsection{Long-term Task Evaluation}
\begin{table}[!t]
\setlength{\abovecaptionskip}{-2pt}
\caption{\textbf{Success rates of long-horizon tasks and their subtasks}, averaged over 20 trials. For entries under ‘Ours’, the first and second values denote success rates under \textit{cached} and \textit{non-cached} conditions, respectively. Compared with DovSG~\cite{dovsg}, our approach consistently yields higher success rates and is capable of completing operations such as door opening, which DovSG~\cite{dovsg} cannot handle.}
\label{5demo}
  
\begin{center}
\setlength{\tabcolsep}{1pt} 
\renewcommand{\arraystretch}{1.2} 

\Large
\resizebox{\linewidth}{!}{
\begin{tabular}{ccccc}
\toprule

\multicolumn{5}{c}{\textbf{ Bring the water bottle from the table by the door and pour it into the bowl.}} \\
 & Open Door(\%) & Pick Bottle(\%) & Pour Water(\%) & Long Term(\%) \\
 DovSG~\cite{dovsg} & \ding{55}  & 75 & \ding{55}  & \ding{55}  \\
 Ours & 60/50 & 85/80 & 70/60 & 35/25 \\
\midrule[\heavyrulewidth] 

\multicolumn{5}{c}{ \textbf{Take the apples from the white box and place them on the cutting board.}} \\
 & Open Box(\%) & Pick Apple(\%) & Place Apple(\%) & Long Term(\%) \\
 DovSG~\cite{dovsg} & \ding{55}  & 50 & 90  & \ding{55}  \\
 Ours & 55/50 & 80/70 & 90/90 & 40/30 \\
\midrule[\heavyrulewidth]

\multicolumn{5}{c}{ \textbf{Place the apples in the drawer and make sure to avoid obstacles.}} \\
 & Open Drawer(\%) & Pick Apple(\%) & Place Apple(\%) & Long Term(\%) \\
 DovSG~\cite{dovsg} & \ding{55}  & 90  & 80  & \ding{55}  \\
 Ours & 85/80 & 95/90 & 90/85 & 70/65 \\
\midrule[\heavyrulewidth]
\multicolumn{5}{c}{ \textbf{Move the tennis ball from the first table to the pink bowl on the third table.}} \\
 & Pick ball(\%) & Place ball(\%) & - & Long Term(\%) \\
 DovSG~\cite{dovsg} & 85  & 90  & - & 75  \\
 Ours & 95/95 & 95/95 & - & 90/90 \\
\midrule[\heavyrulewidth]

\multicolumn{5}{c}{\textbf{  Get a cloth and wipe the stains off the table.}} \\
 & Pick Cloth(\%) & Wipe Table(\%) & - & Long Term(\%) \\
 DovSG~\cite{dovsg} & 60  & \ding{55}  & - & \ding{55}  \\
 Ours & 85/85 & 75/70 & - & 65/60 \\
\bottomrule

\end{tabular}
}
\vspace{-1.5em}
\end{center}
\end{table}

We design five multi-step tasks that couple navigation with contact-rich manipulation: \textbf{1)} Bring the water bottle from the table by the door and pour it into the bowl. \textbf{2)} Pick up apples from the white box and place them on the cutting board. \textbf{3)} Move a tennis ball from the first table to a pink bowl located on the third table. \textbf{4)} Store the apples inside a drawer while avoiding surrounding obstacles. \textbf{5)} Retrieve a cleaning cloth and wipe stains off the table surface. 
Each task is repeated 20 times.

We report results in Table~\ref{5demo} and compare against DovSG~\cite{dovsg}. ReKep~\cite{huang2024rekep} and VoxPoser~\cite{huang2023voxposer} cannot execute long-term procedures end-to-end and are therefore omitted from this setting. 
We evaluate two conditions: \textbf{cached}, where previously generated code for the same task or familiar object can be directly reused, and \textbf{non-cached}, where no prior code is available.

Our method attains comparable success rates between cached and non-cached conditions.
This indicates that EmbodiedCoder does not rely on task-specific templates and that the two-stage pipeline—geometric parameterization followed by constraint-aware trajectory synthesis—supports zero-shot generalization. 
In particular, success rates in the cached setting are slightly higher because execution can proceed by directly reusing code that has already been verified in previous trials, thereby avoiding potential failures caused by incorrect code generated by the coding model.
In contrast, DovSG~\cite{dovsg} performs well on short pick-and-place segments but fails on tasks requiring additional structure, such as door or drawer operation. This gap is consistent with our method’s ability to (i) parameterize articulated and functional parts (e.g., door handle as a cylinder) and (ii) synthesize trajectories constrained by those parameters and by robot kinematics.
Additionally, the door-opening success rate is lower than other subtasks. 
One primary failure mode was observed and is consistent with the method’s assumptions.
When navigating close to the door, the limited field of view of the camera sometimes fails to capture the entire door, resulting in incomplete point clouds. This leads to errors in parameter estimation, particularly in computing the rotation radius, which propagates to trajectory generation and causes execution failure.
These findings highlight the challenges of handling multi-stage tasks under current model capabilities.

\subsection{Simple Task Evaluation}

\begin{table}[h]
\caption{\textbf{Comparison with code-generation methods.} Success rates (\%) are reported using the results of the other methods from their original papers.}

\vspace{-0.5em}
\label{ab_coder}
\centering
\setlength{\tabcolsep}{8pt}
\begin{tabular}{lccc}
\hline
Task (\%) & \text{Pour Tea} & \text{Recycle Can} & \text{Stow Book} \\
\hline
ReKep~\cite{huang2024rekep} & 80 & 80 & 60 \\
VoxPoser~\cite{huang2023voxposer} & 0 & 30 & 0 \\
Code-as-Monitor~\cite{codeasmonitor} & 50 & - & 70 \\
Ours & 80 & 100 & 80 \\
\hline
\end{tabular}
\end{table}
\begin{figure*}[!th]
    \centering
    \includegraphics[width=1\linewidth]{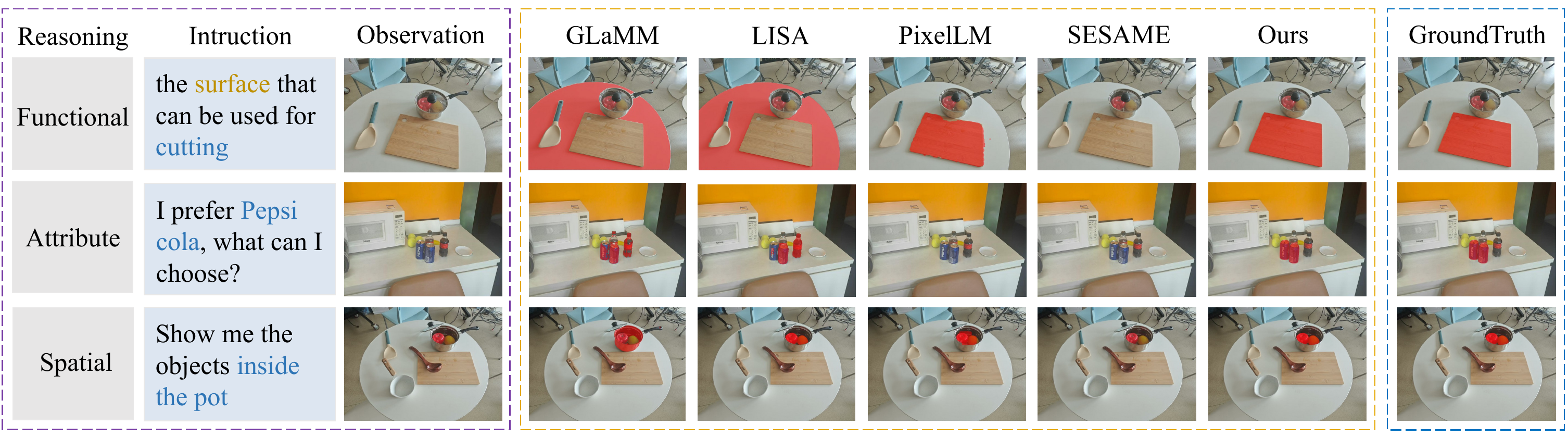}
    \caption{Comparison of different large models on semantic grounding across functional, attribute, and spatial reasoning tasks.}
    \label{fig:segment_results}
    \vspace{-1.0em}
\end{figure*}

\begin{table}[h]
\vspace{-0.5em}
\setlength{\abovecaptionskip}{-5pt}

\caption{\textbf{Quantitative results on simple tasks compared with VLA models.} Success rates (\%) are reported using the results of the other methods from their original papers.}
\label{simpleTask}
\tiny  
\begin{center}
\setlength{\tabcolsep}{1.5pt}
\resizebox{\linewidth}{!}{
\begin{tabular}{ccccccc}

\hline

Task (\%) &  RT-1 & RT-2 & Octo & OpenVLA & RDT & Ours \\
\hline
   \text{Pick Pepsi Can} & 60 & 100 & 0 & 80 & - & 100 \\

\text{Pick Banana} & 100 & 100 & 60 & 100 & - & 80  \\

\text{Pick Green Cup} & 20 & 100 & 0 & 100 & - & 100 \\

\text{Place Apple on Plate} & 0 & 80 & 0 & 80 & - & 95  \\

\text{Place Banana in Pan} & 0 & 40 & 0 & 80 & - & 80  \\

\text{Pour Water} & - & - & 13 & 0 & 63 & 80 \\


\textbf{Average} & 36 & 84 & 12.2 & 73.3 & - & 89.2 \\
\hline

\end{tabular}}
\end{center}
\vspace{-1.5em}
\end{table}

To compare with methods designed for single-step tasks, we evaluate simple tasks drawn from the out-of-distribution benchmarks used in the VLA papers, as shown in Table~\ref{simpleTask}. Our method matches or surpasses trained VLAs without \textbf{additional training}. 
Tasks involving delicate flow control (e.g., pouring) show a clear advantage. This outcome is aligned with the method: the parameterization stage exposes stable surface normals, contact lines, and symmetry axes, and the trajectory code can explicitly encode motion constraints such as tilt angle limits and collision margins. By grounding semantics in the metric scene through parameterized geometry, the system avoids the ambiguity that can arise when control is derived only from learned visual features.

We also compare to code-generation approaches that predict constraint points or action scripts such as ReKep~\cite{huang2024rekep}, VoxPoser~\cite{huang2023voxposer},  and Code-as-Monitor~\cite{codeasmonitor} in Table~\ref{ab_coder}.
Our method attains consistently higher success. 
Unlike other approaches that infer control points or directly output trajectories, EmbodiedCoder first fits parameterized representations, which simplifies following trajectory planning and improves task success rate and stability.

\subsection{Ablation Study}
\subsubsection{Effect of Object Shape}
We assess grasping across objects with distinct morphology (bottles, oranges, plastic bags) and compare with AnyGrasp~\cite{anyGrasp} (Table~\ref{ablation_anygrasp}).
\begin{table}[h]
\caption{Comparison with AnyGrasp~\cite{anyGrasp} on grasp success rates (\%) across different objects over 20 trials.}
\vspace{-1em}
\label{ablation_anygrasp}
\tiny
\begin{center}

\setlength{\tabcolsep}{1.5pt}

\resizebox{\linewidth}{!}{

\begin{tabular}{cccccccc}

\hline

\multirow{2}{*}{Task (\%)} & \multirow{2}{*}{Bottle} & \multirow{2}{*}{Apple} & \multirow{2}{*}{Orange} & \multirow{2}{*}{Banana} & Pepsi & Plastic & Green \\
& & & & & Can & Bag & Cup \\
\hline
\text{Anygrasp~\cite{anyGrasp}} & 95 & 70 & 95 & 80 & 40 & 90 & 60 \\

\text{Ours} & 100 & 95 & 100 & 90 & 75 & 100 & 80 \\

\hline

\end{tabular}
}
\end{center}
\vspace{-1.5em}
\end{table}
Our method achieves higher success rates. The improvement aligns with our parameterization design: for spheres, the planner aligns the gripper along the radial direction inferred from the fitted center and radius; for cylinders, it selects contact along the principal axis while respecting gripper aperture. AnyGrasp predicts grasps directly on point clouds without encoding gripper and kinematic constraints, which can yield unreachable or unstable poses on our platform.

\subsubsection{Robustness of Semantic Grounding}
We qualitatively assess semantic grounding across three categories of reasoning: functional, attribute, and spatial, as illustrated in Fig.~\ref{fig:segment_results}. The visualizations reveal that different models show considerable variation in their reasoning performance. Reliable grounding and accurate segmentation of task-relevant objects are essential, since errors in location or size can directly compromise subsequent geometric parameterization and trajectory execution.

We further examined the effect of providing a two-dimensional semantic map as input to the VLM. As shown in Table~\ref{compare}, with this representation, the model produced more feasible task decompositions, such as navigating to a doorway before opening it for cross-room tasks. The results indicate that a 2D semantic map helps align task decomposition with real environments and reduces planning hallucinations.

\begin{table}[h]

\caption{Comparison of five different VLMs on subtask decomposition success rates (\%) with and without map.}
\vspace{-1.1em}
\begin{center}
\setlength{\tabcolsep}{1.5pt}
\resizebox{\linewidth}{!}{
\begin{tabular}{lcccccc}
\hline
Models & PaliGemma & Qwen-3B & Qwen-7B & GPT-5 & Gemini2.5-Pro \\
\hline
w/. Map    & 0 & 80 & 88  & 88 & 56 \\
w/o. Map     & 0 & 72 & 72  & 64 & 20 \\
\hline
\end{tabular}}
\end{center}
\vspace{-2em}
\label{compare}
\end{table}

\begin{figure}[h]
    \centering
    \includegraphics[width=1\linewidth]{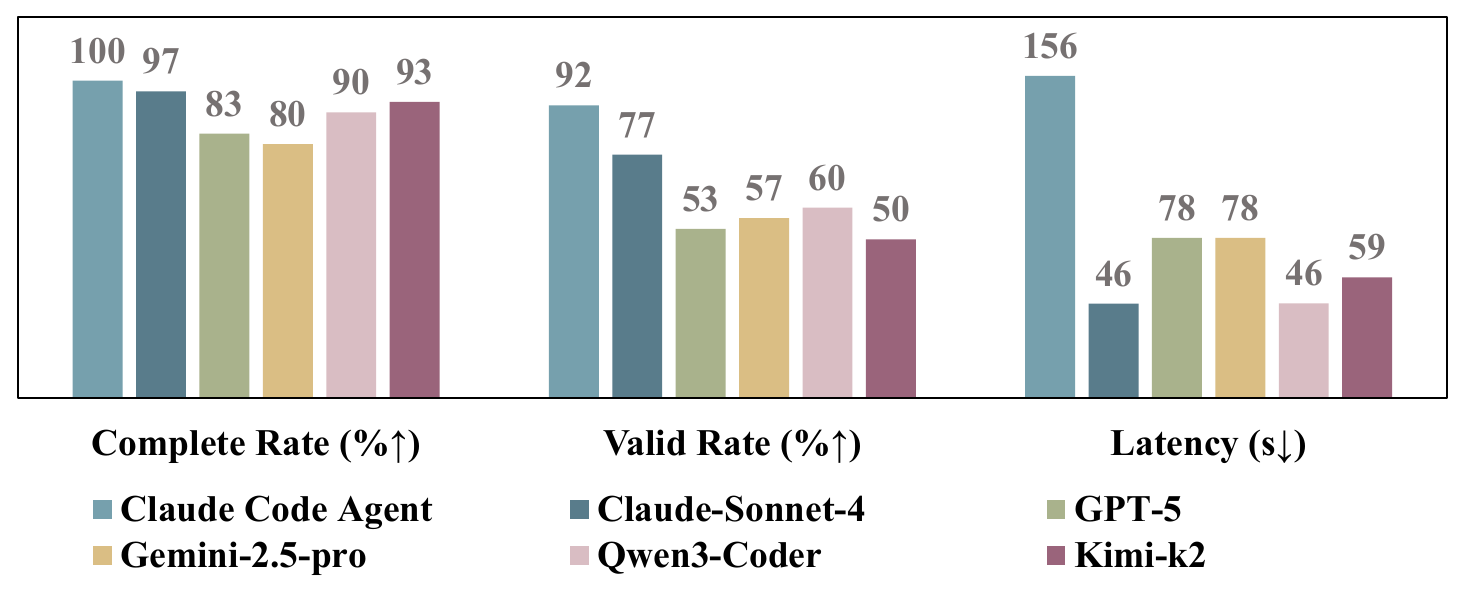}
    \caption{Comparison of different coding models in their ability to perform geometric parameterization and trajectory synthesis under the same prompt.}
    \label{fig:Coding Capabilities}
    \vspace{-1.0em}
\end{figure}
    
\subsubsection{Impact of Coding Models Capabilities}
As shown in Fig.~\ref{fig:Coding Capabilities}, we compare different coding models on their success rates in object geometric parameterization and task-oriented trajectory synthesis. The complete rate measures whether the model can generate executable code, while the valid rate measures whether the generated code successfully accomplishes the intended task, which includes both parameterization and trajectory synthesis.
Claude-Sonnet-4~\cite{claude} achieves the highest success rates but also exhibits the largest latency. Other models perform significantly worse, indicating that only recent coding models possess sufficient reasoning ability for these tasks. Consequently, our paradigm is feasible only when coding models are strong enough to support reliable task reasoning and execution.
\section{CONCLUSION AND LIMITATIONS}
We introduce EmbodiedCoder, a training-free framework that combines large language and coding models with structured geometric object representations to enable open-world mobile manipulation. By grounding semantic knowledge into parameterized affordances and generating executable trajectory code, the system allows robots to perform long-term, contact-rich tasks without predefined primitives or additional training. Experiments on real robots demonstrate strong generalization to novel objects and environments, advancing the integration of high-level reasoning with low-level control. Nonetheless, several limitations remain. First, task success is highly sensitive to the quality of code generated by large models, and errors in logic or syntax can significantly reduce reliability. Second, the code synthesis process introduces latency, which may limit responsiveness in real-time applications. Addressing these issues will enhance robustness and scalability, moving closer to practical deployment of versatile robot intelligence.

\bibliographystyle{IEEEtran}
\bibliography{root}

\end{document}